\documentclass{article}

\usepackage[final, nonatbib]{neurips_2023}

\usepackage[utf8]{inputenc} %
\usepackage[T1]{fontenc}    %
\usepackage{hyperref}       %
\usepackage{url}            %
\usepackage{booktabs}       %
\usepackage{amsfonts}       %
\usepackage{nicefrac}       %
\usepackage{microtype}      %
\usepackage{xcolor}         %
\usepackage{graphicx}
\usepackage{caption}
\usepackage{subcaption}

\title{UniTeam: Open Vocabulary Mobile Manipulation Challenge}

\author{%
  Andrew~Melnik\\
  Bielefeld University, Germany\\
  \texttt{andrew.melnik.papers@gmail.com} \\
  \And
  Michael Büttner\\
  Bielefeld University, Germany\\
  \texttt{mbuettner@techfak.uni-bielefeld.de} \\
  \And
  Leon Harz\\
  Bielefeld University, Germany\\
  \And
  Lyon Brown\\
  Bielefeld University, Germany\\
  \And
  Gora Chand Nandi\\
  IIIT-A, India\\
  \And
  Arjun PS\\
  IIIT-A, India\\
  \And
  Gaurav Kumar Yadav\\
  IIIT-A, India\\
  \And
  Rahul Kala\\
  ABV-IIIT-M, India\\
  \And
  Robert Haschke\\
  Bielefeld University, Germany\\
}  

\begin{document}

\maketitle

\begin{abstract}
This report introduces our \textit{UniTeam} agent - an improved baseline for the \textit{HomeRobot: Open Vocabulary Mobile Manipulation} challenge. The challenge poses problems of navigation in unfamiliar environments, manipulation of novel objects, and recognition of open-vocabulary object classes. This challenge aims to facilitate cross-cutting research in embodied AI using recent advances in machine learning, computer vision, natural language, and robotics. In this work, we conducted an exhaustive evaluation of the provided baseline agent; identified deficiencies in perception, navigation, and manipulation skills; and improved the \textit{baseline agent}'s performance. Notably, enhancements were made in perception - minimizing misclassifications; navigation - preventing infinite loop commitments; picking - addressing failures due to changing object visibility; and placing - ensuring accurate positioning for successful object placement.

\end{abstract}

\begin{figure}
  \centering
     \begin{subfigure}[b]{1\textwidth}
         \centering
         \includegraphics[width=\textwidth]{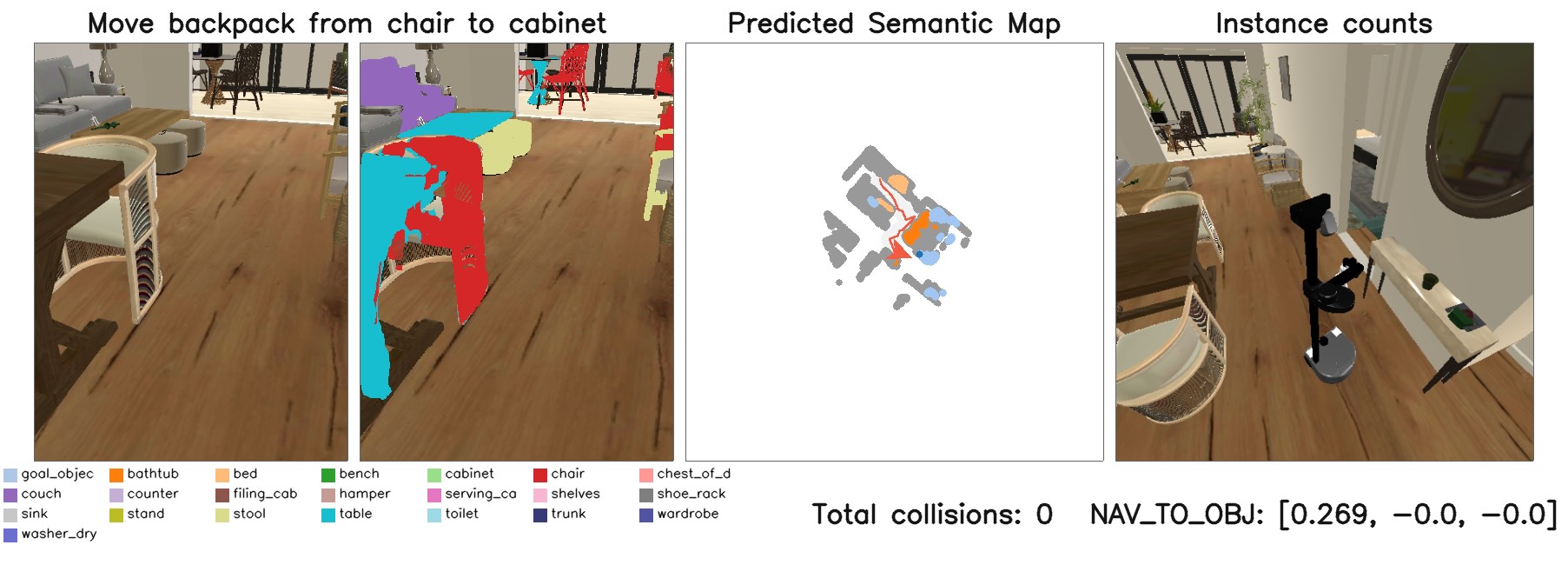}
         \caption{\textit{Find object}-phase. In this phase, the agent searches for the backpack. In the predicted semantic map, the blue clusters represent unexplored chairs, the dark orange clusters represent explored chairs and the light orange clusters represent cabinets.}
         \label{fig:panel_nav_to_obj}
     \end{subfigure}
     \begin{subfigure}[b]{1\textwidth}
         \centering
         \includegraphics[width=\textwidth]{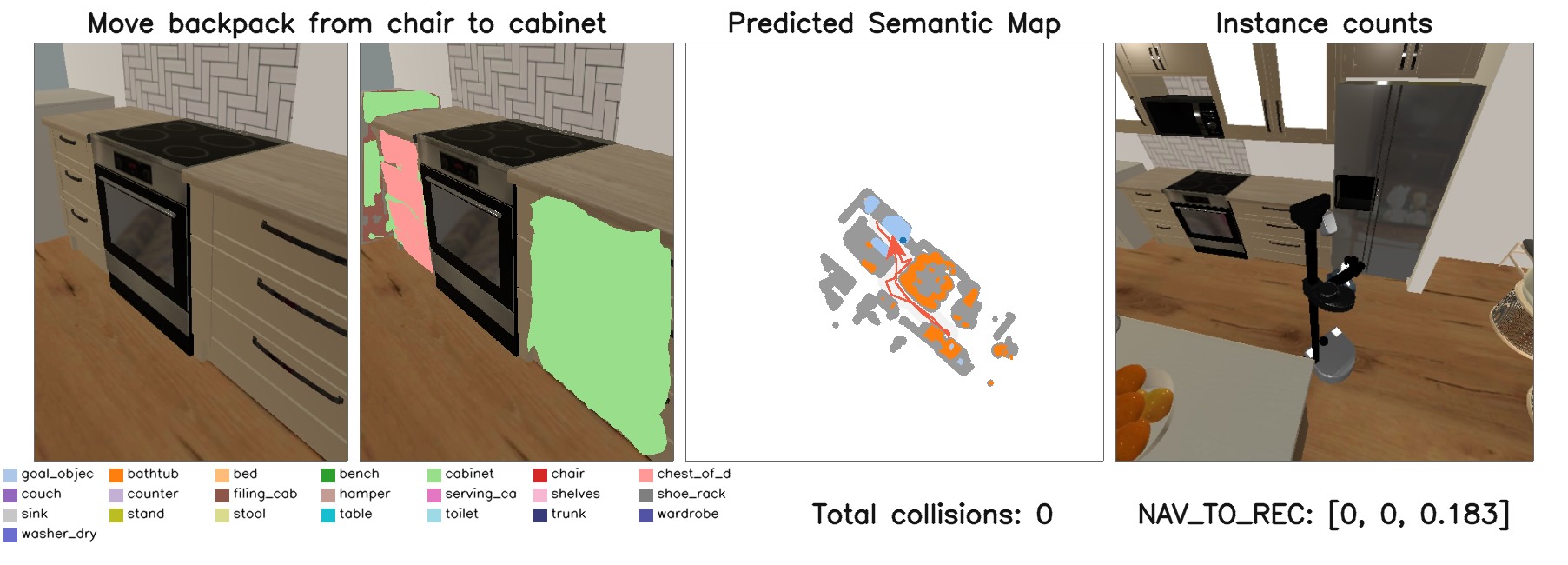}
         \caption{\textit{Navigate to end receptacle}-phase. In this phase, the agent moves towards a cabinet to place the backpack on. In the semantic map, the blue clusters represent cabinets and the orange clusters represent chairs.}
         \label{fig:panel_nav_to_rec}
     \end{subfigure}
  \caption{\label{panel} Main panel at two different time steps. The instruction in this episode is "Move backpack from chair to cabinet".  From left to right you can see: The RGB image from the agent's camera; the segmentation image by Detic; the predicted semantic map; a third person view of the agent (just for visualization purposes). The light blue clusters on the predicted semantic map represent all goal points, while the dark blue point represent the goal point to which the agent has chosen to navigate towards. On the bottom you can see a legend for the segmentations, the total amount of collisions, the current phase and the next action the agent is taking.}
\end{figure}

\section{Baseline}

The challenge sets a Stretch robot~\cite{kemp2022design} in simulated and real human indoor environments with the goal of \textbf{moving the object from the start receptacle to the end receptacle}. The receptacles, which are classes of furniture or fixtures, don't have to be a single entity in the environment, for example if the end receptacle is "table", there are usually multiple tables in the apartment of which any of them is correct. The HomeRobot Framework~\cite{yenamandra2023homerobot} includes two primary baseline approaches: a heuristic baseline utilizing well-established motion planning techniques alongside simple rules for executing grasping and manipulation actions; and an reinforcement learning baseline, where exploration and manipulation skills are acquired using the off-the-shelf policy learning algorithm DD-PPO \cite{bach2020learn}. In this work, we will discuss the heuristic baseline and our improvements on it.

The baseline agent's capabilities are divided into the following key skills:
\begin{enumerate}
    \item \textbf{Detection}: object detection and segmentation on the agent's camera RGBD images. Building \textit{Bird's Eye View} (BEV) map with semantic areas.
    \item \textbf{Exploration}: Exploring the environment to find \texttt{start\_receptacles} and \texttt{end\_receptacle}. Exploring \texttt{start\_receptacles} to find the object.
    \item \textbf{Navigation}: Move towards the specified navigation goal: the object, start- or end receptacle.
    \item \textbf{Picking}: Picking up the object, supported by a high-level action command due to the absence of simulated gripper interaction in Habitat. The agent is equipped with multiple grasping strategies and supports policies for learning grasping.
    \item \textbf{Placing}: Adjusting the agent's position in front of the \texttt{end\_receptacle} and placing the object.
\end{enumerate}

The skills are executed in a certain order. At the beginning, the agent turns 360° to get an overview of its surroundings. After that, the agent uses the Exploration skill to find the object, navigates towards it if the object is found and uses the Picking skill to pick it up. If no entity of the end receptacle is found, the agent uses the Exploration skill again, otherwise it uses Navigation skill to move to the nearest end receptacle and Placing skill to place the object. 
The Detection skill is executed in every frame of the run. This means that the agent collects information of start and end receptacles even if the current goal is to find the object. The baseline uses the perception model Detic~\cite{zhou2022detecting} to generate masks for objects and receptacles.
A panel showing aspects of the baseline approach during a run can be found in Figure~\ref{panel}.

\begin{figure}
  \centering
  \includegraphics[width=1.0\textwidth]{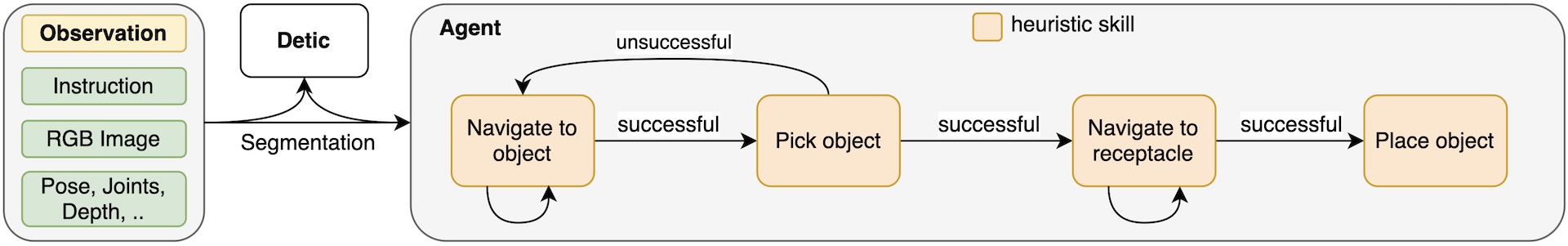}
  \caption{\label{architecture} The architecture of our agent. It is build as a state machine with six states: \textit{Find object}, \textit{Navigate to object}, \textit{Pick object}, \textit{Find end receptacle}, \textit{Navigate to end receptacle}, and \textit{Place object}. In the \textit{Find}-phases, the \textit{Exploration} skill is used. The agent's perception model is Detic \cite{zhou2022detecting}, it segments the object and receptacles in the RGB image. This segmentation, together with other observations such as the pose, joints and a depth image, are the input to the agent.}
\end{figure}

\newpage
\section{Method}
An overview of our \textit{UniTeam}-agent's architecture can be found in Figure~\ref{architecture}.

\subsection{Detection}
Initially, the baseline agent employed the same confidence threshold value for all receptacles and objects. During testing, it became evident that the agent frequently overlooked objects with this threshold value. However, reducing the threshold value increased the misclassification rate of \textit{end receptacles}. The implementation of a dynamic object/receptacle-specific threshold proved effective in mitigating this issue. 

\begin{figure}
  \centering
  \includegraphics[width=0.5\textwidth]{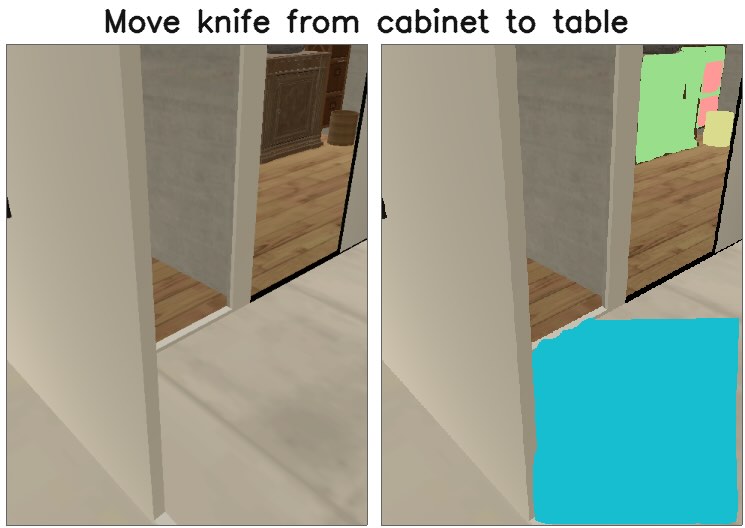}
  \caption{\label{floor} Example for incorrect floor classifications. The floor is classified as a table (blue cluster), which is the \textit{end receptacle}. If the floor wasn't filtered out, the agent would want to place the knife on there.}
\end{figure}

The perception module also encounters challenges due to incorrect detection of
objects on the floor (see Fig.~\ref{floor}) or the floor itself as objects or receptacles. To minimize the occurrence of these false positives \cite{melnik2021critic}, we established a height threshold for receptacles within the perception module using the depth information. This eliminates detections corresponding to the floor level, thereby facilitating more accurate planning.

\subsection{Exploration}
\begin{figure}
  \centering
  \includegraphics[width=0.5\textwidth]{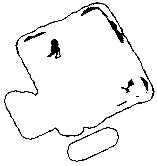}
  \caption{\label{goal_map} The agent's frontier map shown as the larger enclosed shape, and the amalgamation of exploration goal points inside. The exploration goal points adjacent to identified \textit{start receptacles} indicate free space for exploring. The intention of the exploration is to check from different sides whether the goal object is located on this receptacle.
  }
\end{figure}

The baseline agent frequently encountered challenges when strictly adhering to a single goal, as the heuristic consistently selected the same action, leading to infinite oscillations between two adjacent states. To mitigate this issue, a decision-making mechanism was introduced to operate on the goal map, as depicted in Figure~\ref{goal_map}. This mechanism allows the agent to choose between the goal point and exploring the frontier during navigation, preventing persistent commitment to an unattainable goal.

In the baseline approach, the agent directs itself towards the nearest identified receptacle. However, due to imperfect object detection, the nearest receptacle may not align with the intended goal. To address this, our agent stores the probabilities of detected receptacles in the Bird's Eye View (BEV) map \cite{harter2020solving}. Subsequently, the agent selects the reachable receptacle with the highest matching probability, reducing the likelihood of placing the object on incorrectly classified receptacles. As the agent begins adding identified \textit{end receptacles} to the goal map during the \textit{Navigate to object} phase, a sufficient number of \textit{end receptacles} are typically available for selection.

\subsection{Navigation}
 When the baseline agent detected a potential collision, the collision was not incorporated efficiently into the new planning, leading the agent to attempt the same movement as before. To rectify this behavior, the detected collision is now integrated at the last short-term goal on the obstacle map. This refinement provides a more detailed representation of the environment in the BEV map, enabling the agent to formulate improved plans. Due to the strategy followed by the baseline agent, instances arose where the agent was not well-aligned with the \textit{end receptacle}, resulting in an inadequate position for successful object placement. To address this limitation, the navigation in our agent aims to position the agent near the center of the goal receptacles. This adjustment ensures generally improved starting positions for the object placement routine.

\subsection{Picking}
The baseline picking routine necessitates the presence of the object of interest within the field of view of the agent's camera. However, in certain scenarios, the object is not visible anymore after the agent moved towards the receptacle, thus resulting in picking failures. To address this issue, out agent is rotates left and right if the initial picking attempt fails, expanding its overall field of view of the receptacle and thereby increasing the probability of object detection.

In our agent a check has been introduced to verify whether the object was successfully picked up. Only upon a successful pick, the agent transitions to the next phase; in case of a failed pick, however, our agent reverts to the navigation phase, exploring different locations for the object.

\subsection{Placing}
The baseline agent encountered challenges in accurately calculating the distance between the agent and the placing point, resulting in attempts to move either too far away or too close to the receptacle. This issue was mitigated in our agent by moving the agent toward the receptacle in small steps over multiple iterations, ensuring that the agent is positioned appropriately for successful object placement.

Additionally, the baseline agent often placed objects close to the edges of receptacles, causing the object to fall down. To address this, the placing point was adjusted to maintain a safe distance from the receptacle edges, preventing later object falling down.

In situations where the end receptacle is not recognized during the placing phase, the baseline agent previously placed the object blindly, resulting in failures due to collisions or misplaced objects. To mitigate this, our agent places the object on a surface if the end receptacle is no longer detected, reducing the occurrence of misplacements.

Furthermore, in the placing phase, large objects grasped by the baseline agent frequently collide with the \textit{end receptacle}. This issue is alleviated in our agent by dropping the object onto the \textit{end receptacle} from an increased height.

\section{Evaluation}

The HomeRobot OVMM Challenge comprises two primary evaluation stages~\cite{OVMMChallenge2023}:

\begin{enumerate}
    \item \textbf{Simulation Phase:} Systems are first evaluated in a simulated environment, replicating real-world home settings with multiple rooms and varying layouts. This phase is crucial for testing the agent's ability to navigate and manipulate objects in a controlled yet diverse setting.
    \item \textbf{Physical Robot Phase:} Agents are tested on a physical robot in a real-world environment, featuring novel objects and layouts. This phase assesses the agent's adaptability and performance in real-world conditions.
\end{enumerate}

\subsection{Simulation Phase}
The simulation evaluation encompasses the following aspects~\cite{OVMMChallenge2023}:

\begin{itemize}
    \item \textbf{Environment Setup:} The simulation involves 50 scenes, each offering unique challenges. These scenes are designed to test the agent's interaction with a variety of objects, including both seen and unseen categories.
    \item \textbf{Computational Resources:} Evaluations are conducted on AWS EC2 p2.xlarge instances, which include Tesla K80 GPUs, 4 CPU cores, and 61 GB RAM. This setup ensures a robust platform for performance assessment .
    \item \textbf{Testing Protocol:} Agents are tested over 1000 episodes, providing a comprehensive evaluation of their capabilities in diverse scenarios.
\end{itemize}

\subsection{Real-World Evaluation}
Key aspects of the real-world evaluation include~\cite{OVMMChallenge2023}:

\begin{itemize}
    \item \textbf{Evaluation Environment:} The physical robot testing occurs in a fully equipped apartment with novel objects and furniture layouts, hosted by Meta in Fremont, California \cite{OVMMChallenge2023}.
    \item \textbf{Agent Adaptability:} This phase particularly focuses on the agent's ability to handle real-world dynamics and novel scenarios, providing a critical assessment of its practical applicability.
\end{itemize}

\subsection{Evaluation Metrics}
The challenge employs a set of metrics to assess the agents' performance:

\begin{enumerate}
    \item \textbf{Overall Success:} correct placement of the object on the target receptacle.
    \item \textbf{Partial Success:} completion of individual sub-tasks within each episode.
    \item \textbf{Number of Steps:} number of steps required to complete an episode.
\end{enumerate}

\subsection{Phases and Splits}
The challenge is divided into several phases, each designed to test different aspects of the agents' capabilities:

\begin{enumerate}
    \item \textbf{Minival Phase:} A preliminary phase for sanity checking.
    \item \textbf{Test Standard Phase:} Serves as the public leaderboard and basis for performance reporting.
    \item \textbf{Test Challenge Phase:} Determines the finalists of the Simulation Phase to proceed to the physical robot evaluation.
\end{enumerate}

\subsection{Results}

\begin{table}
\centering
{%
\begin{tabular}{@{}llll@{}}
\toprule
\multicolumn{4}{c}{\textbf{Results of Test Standard Phase \cite{homerobotleaderboard}}}                                             \\ \midrule
\multicolumn{1}{l}{\textbf{Agent name}} &
  \multicolumn{1}{c}{\textbf{Overall Success}} &
  \multicolumn{1}{c}{\textbf{Partial Success}} &
  \multicolumn{1}{c}{\textbf{Number of Steps}} \\ \midrule
\begin{tabular}[c]{@{}l@{}}KuzHum\end{tabular} &
  2.8\% &
  19.1\% &
  1152.99 \\ \midrule
UniTeam (our agent)      & \textbf{2\%}           & \textbf{18.6\%}          & \textbf{1140.54}         \\ \midrule
PieSquare & 2\%           & 11.1\%          & 1082.56         \\ \midrule
Baseline (heurisitc)    & 0\% & 10.5\% & 1063.74 \\ \bottomrule
\end{tabular}%
}
\vspace{0.1in}
\caption{Evaluation results of the top three agents over the 1000 episodes of the \textit{Test Standard Phase}, in comparison to the heuristic baseline.}
\label{tab:result}
\end{table}

Through incremental enhancements to the baseline agent, as described above, performance steadily improved, culminating in a score of 2\% overall success during the Test Standard Phase (see Table \ref{tab:result}). In comparison to the heuristic and reinforcement-learned baselines, the our improved baseline agent exhibits notable performance gains through described adjustments. Notably, refining the perception module had the most substantial impact on overall performance.

\section{Conclusion}

We implemented improvements on several aspects of the baseline agent. Utilizing these improvements, we achieved an overall success score of 2\% and a partial success score of 18.6\% in the simulation phase of the Open Vocabulary Mobile Manipulation challenge. This challenge aims to facilitate cross-cutting research in embodied AI using recent advances in machine learning, computer vision, natural language, and robotics \cite{rana2023contrastive}.

Our analysis revealed substantial areas for further improvement. Throughout the simulation, we encountered instances where the embodied agent \cite{konig2018embodied} struggled with challenges such as attempting to place an object on a receptacle that is too large, attempting to place an object through a wall, getting stuck or abandoning navigation in narrow passages, overlooking the object, or timing out due to difficulty locating the object, etc. We found that correctly recognizing the object and the receptacles is still the primary bottleneck, serving as the root cause for the majority of unsuccessful episodes.

To further enhance the mobile manipulation performance of the agent, one could explore Behavior Search methods \cite{beohar2022planning, malato2022behavioral, malato2023behavioral, sheikh2023language}, which integrate  few-shot learning and Imitation Learning advancements, leveraging demonstration datasets. While training another perception model carries inherent costs, implementing ideas of modular end-to-end learning \cite{melnik2019modularization} may mitigate some of those problems. However, our approach of filtering the input prompts of Detic~\cite{zhou2022detecting} or GroundedSAM~\cite{liu2023grounding, kirillov2023segany}, an alternative segmentation model, by using Recognise Anything~\cite{zhang2023recognize} did not lead to the desired improvements. Using a concept graph~\cite{gu2023conceptgraphs} may solve the problem of classifying the same receptacle differently in consecutive frames. Additionally, addressing the other identified challenges by implementing heuristic solutions emerges as a pragmatic and stable avenue for advancing our agent. 

{\small
\bibliography{bibliography.bib}
\bibliographystyle{ieeetr}
}

\end{document}